\documentclass{svproc}
\usepackage{url}

\usepackage{graphicx}
\usepackage{microtype}
\usepackage{flushend}
\usepackage{adjustbox}
\usepackage[dvipsnames]{xcolor}
\usepackage{hyperref}
\usepackage{authblk}
\hypersetup{
	colorlinks=true,
	linkcolor=blue,
	filecolor=blue,      
	urlcolor=blue,
	citecolor = blue
}
\usepackage[caption=false]{subfig}
\usepackage{tabularx}
\newcommand{\WordTwoVec}{Word\-2Vec }

\begin{document}
\bibliographystyle{spmpsci}
\mainmatter              
\title{Seq2Seq and Joint Learning Based Unix Command Line Prediction System}



%
\titlerunning{Seq2Seq and Joint Learning Based Unix Command Line Prediction System}  

\author{Thoudam Doren Singh\inst{1} \and Abdullah Faiz Ur Rahman Khilji\inst{1} \and
Divyansha\inst{1} \and Apoorva Vikram Singh\inst{2} \and Surmila Thokchom\inst{3} \and Sivaji Bandyopadhyay\inst{1}}

\authorrunning{TD Singh, AF Khilji, Divyansha, AV Singh, S Thokchom, Sivaji B.  et al.} 
%
%
\institute{Center for Natural Language Processing (CNLP) and Department of Computer Science and Engineering, National  Institute of Technology Silchar, Assam, India\\
\and
Department of Electrical Engineering, National  Institute of Technology Silchar, Assam, India\\
\and
Department of Computer Science and Engineering, National  Institute of Technology Meghalaya, Shillong, India\\
\email{\{thoudam.doren, abdullahkhilji.nits, divyansha1115, singhapoorva388, surmila.th, sivaji.cse.ju\}@gmail.com}}

\maketitle              
	\begin{abstract}
		Despite being an open-source operating system pioneered in the early ’90s, UNIX based platforms have not been able to garner an overwhelming reception from amateur end users. One of the rationales for under popularity of UNIX based systems is the steep learning curve corresponding to them due to extensive use of command line interface instead of usual interactive graphical user interface. In past years, the majority of insights used to explore the concern are eminently centered around the notion of utilizing chronic log history of the user to make the prediction of successive command. The approaches directed at anatomization of this notion are predominantly in accordance with Probabilistic inference models. The techniques employed in past, however, have not been competent enough to address the predicament as legitimately as anticipated. Instead of deploying usual mechanism of recommendation systems, we have employed a simple yet novel approach of Seq2seq model by leveraging continuous representations of self-curated exhaustive Knowledge Base (KB) to enhance the embedding employed in the model. This work describes an assistive, adaptive and dynamic way of enhancing UNIX command line prediction systems. Experimental methods state that our model has achieved accuracy surpassing mixture of other techniques and adaptive command line interface mechanism as acclaimed in the past.
	\end{abstract}

\section{Introduction}
\label{introduction}
The work aims at resolving the long-standing plight of unfamiliarity with command line interface in UNIX based systems. This will not only improve the efficiency of the user but also improve the learning curve for the beginners. The concerned research work treats the problem of UNIX Command Line Prediction as a \textit{sequence prediction problem} instead of the traditionally adapted provision of recommendation systems. RNN \cite{mikolov2010recurrent} is able to “theoretically” use information from the past in predicting the future. However, plain RNNs suffer from vanishing and exploding gradients problem making them hard to use practically. For this problem, we used LSTM which uses gates to flow gradients back in time solving the vanishing and exploding gradient problem. Thus, with the advent of a command line prediction system involving accurate prediction, a GUI prototype of the UNIX shell can be brought in place subsequently realising the serious necessity of a user-friendly environment for amateur end users. Our model delves into the \textit{user’s bash history} and learns from it while providing a thorough path to him/her from the past usage patterns of the professionals and scientists. We have also been able to establish a novel method and outperform the 50\% threshold of accuracy set by previous works. The former maximum accuracy was accomplished by \cite{korvemaker2000predicting} in which \cite{davison1998predicting} was extended to employ consideration of error output with the history of commands and dynamic file name replacement, attaining 47.9\% accuracy.

\section{Related  Work}
\label{related_previous_work}
Investigating ways to customize command line interface has not been a very common practice in the past years. In the past, the relevant work \cite{korvemaker2000predicting} discussed three main algorithms to this end. It comprised of \textit{C4.5}: a well used decision tree algorithm, \textit{Most Recent Command (MRC)} that predicts the same command that was just executed and \textit{Most Frequent Command (MFC)} which predicts the command that previously occurred most frequently in the user’s interactions. The work reported a maximum accuracy of 45\% employing a novel use of extracting feature from the frequency of graph-based induction model and use of data dependency. Another attempt \cite{durant2002predicting} also gives an example of a system in which they had extended \cite{davison1998predicting} while considering error output with the history of commands coupling it with dynamic file name replacement. The maximum accuracy obtained was 47.9\%. Our work differs from the above works on mainly two factors. We used a Seq2seq LSTM model to predict the sequence whose accuracy advances persistently with the amount of data it gathers from the user (since it’s a Deep Learning approach). And secondly, we have also included the KB along with a corpus to ensure more comprehensive embeddings for our model. Using KB and corpus to enhance embeddings is an intuitive idea which has captured research attention recently by the work \cite{alsuhaibani2018jointly}. The table \ref{table1} shows the accuracy of various prediction systems on this domain.

	\begin{table*}
	\centering
	\begin{tabularx}{\textwidth}{lXr}
		\hline\noalign{\smallskip}
		S No.\hspace*{0.5cm} & Paper Name\hspace*{0.5cm} & Accuracy  \\
		\noalign{\smallskip}\hline\noalign{\smallskip}
		1 & Predicting UNIX Command Lines:
		Adjusting to User Patterns, \cite{korvemaker2000predicting} & 47.9\% \\
		2 & Toward An Adaptive Command Line Interface, \cite{davison1997toward}& 45\% \\
		3 & Predicting UNIX commands using decision tables and decision trees, \cite{durant2002predicting}& 42\%\\
		4 & Predicting Sequences of User Actions, \cite{davison1998predicting}& 39.9\% \\
		5 & The Learning Shell, \cite{jacobs2000learning}& 36.4\% \\
		6 & Experiments in UNIX command prediction, \cite{davison1997experiments}& 31\% \\
		7 & User Command Prediction by Graph-Based Induction, \cite{yoshida1994user}& 22\% \\
		\noalign{\smallskip}\hline
	\end{tabularx}
	\caption{Accuracy comparison of previous such works.}
	\label{table1} 
\end{table*}


\section{The methodology of acquiring data}

\subsection{Dataset}
\label{dataset}
We manoeuvered our experiments on “Unix Data Description dataset by Saul Greenberg” \cite{greenberg1988using} which comprises of 168 trace files collected from 168 different users. However, for experimentation enterprise, we utilized the data curated from the group of \textit{52 Computer Scientists}. We substantially preferred this group as tasks performed by this group were least monotonous and more heterogeneous than rest of the user groups making them more turbulent to predict. Since this dataset included many other minute details a comprehensive exercise of data pre-processing was necessary to make the data consistent. The command lines rendering any grade of error were exterminated from the curated dataset. The instances containing any directories/files after a command were substituted with the keyword \textit{“filename”} while occurrences specifying any class of parameters were substituted with the keyword \textit{“parameter”}. The aliases were replaced with the original commands.

\section{Our Novelty}
\subsection{Overview}
	Our main concept for the complete project revolved upon two main facets of our theory. One was employing a highly promising Seq2seq model to cater to the sequence prediction needs of our project while the other contemplated on the KB aspect more or less. We formulated a way to capture the domain knowledge of UNIX commands focusing on commands which had a rare occurrence in the log of Intermediate user and had a little more frequent appearances in those of Specialized Scientists. To accomplish this end, we used KB and have analysed the use of \WordTwoVec \cite{goldberg2014word2vec} and GLoVe \cite{pennington2014glove}.

\subsection{Knowledge Base}
For constructing an exhaustive KB using a list of vocabulary extracted from the dataset, we designed an algorithm friendly to hyperthreading involving an end to end approach to cater to the efficiency requirements of the dataset involving 38,052 command elements. The data scrapped from \url{linux.die.net} was concatenated, all special characters were removed and the resulting data was stemmed to obtain a uniform data for all the commands. For further removing bias due to large data present in some of the commands, common words having 3 characters or less were removed to finally obtain a data ready for \textit{synonym pair extraction}. Since the Knowledge Base (KB) required for the Joint Learning was in synonym form it required that we generate an exhaustive list of command synonyms. Thus iteratively, 5 most similar commands to every command were inserted into the list. The threshold was calculated through probabilistic inference considering and analyzing the mean, median and modes of the number of characters available in the data.
\\




\subsection{Our Approach}
\label{our_approach}
\textit{KB} based word vector embeddings and \textit{Corpus-based} word vector embeddings are two prevalent stratagems deployed to tackle the word embeddings creation scheme. Although, learning word embeddings solely from corpus is a widely practised approach, it has several limitations. KB explicitly defines significance of the words by taking into account the association between words. They include the meaning of those words using semantic relations like synonymy, hypernymy, meronymy, etc. For a KB, every specific word has a finite number of entries unlike a corpus where several co-occurrences between two words can prevail in dissimilar contexts. Due to this reason, the precise estimation of tenacity of relation between words using KB is not attainable. In a nutshell, the drawback of using corpus-based approach is the omission of rich semantic relations while that of KB is the exclusion of contextual information. These two approaches \textit{complement each other.} In other words, they deliver on the limitations of each other. We have utilised GloVe for the purpose of creating enhanced embeddings by jointly learning through KB and corpora, preferring it over \WordTwoVec approach since GloVe based embeddings have outperformed \WordTwoVec based ones as seen in experimental results in section \ref{result}.

\begin{figure*}
	\includegraphics[width=\textwidth]{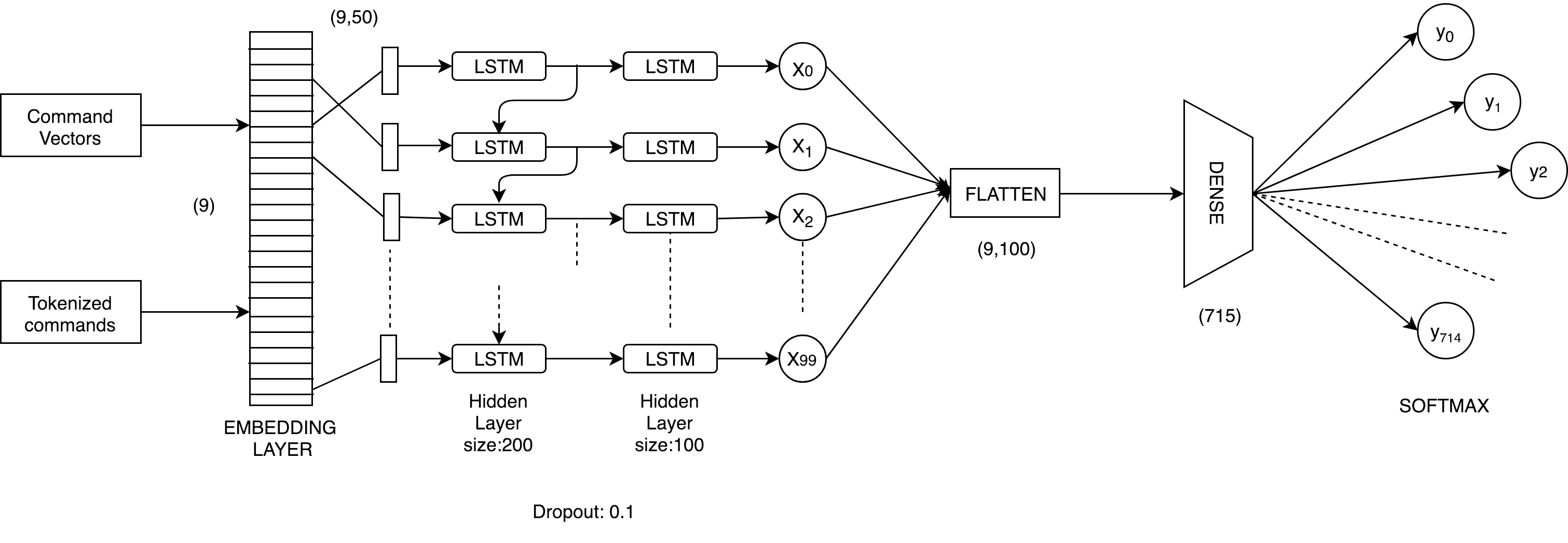}
	\caption{The Model}
	\label{figure4}     
\end{figure*}

\section{A Basic Outline of the model}
For a small-sized data sample along with a high number of covariates \cite{daee2017knowledge}, the “small n, large p” problem, prediction becomes challenging. A standard concern for models that strive to make the prediction for a specific user community by utilising user’s data is the issue of inherent \textit{data sparsity}. A standard user is liable to generate only a limited quantity of data, despite the massive size of the corpus. To address the subject of inherent data sparsity, various ingenious techniques have been devised. The approaches undertaken in the past predominantly incorporated \textit{Probabilistic inference models} which have not turned out to be hugely captivating. The major shortcoming faced when this approach is implemented is its \textit{inability to appreciate the sequential intelligence of the data}. The probabilistic inference method fundamentally makes the prediction by utilizing the number of occurrences of commands in records without acknowledging the sequence in which they have appeared. \textit{Recurrent neural network (RNNs)} architectures offer a fix to this dilemma. Chain-like nature of recurrent neural network (RNNs) reveals that they are intimately connected to list and sequences. This inherent architecture of recurrent neural networks (RNNs) makes it an optimal instrument to deal with our sequential data.

\section{Experimentation}





\begin{table}

\begin{adjustbox}{width=1\textwidth}
    \small
	\centering

	\begin{tabular}{ccccc}
		\textbf{Batch Size}\hspace*{0.5cm} & \textbf{Optimal Epoch}\hspace*{0.5cm} & \textbf{Accuracy on Test Set}\hspace*{0.5cm} & \textbf{10-fold Cross Validation}\hspace*{0.5cm} & \textbf{5-fold Cross Validation}\hspace*{0.5cm} \\
		\hline
		0.3                 & 3                      & 51.43                     & 48.18                                 & \textbf{47.41}                                \\
		0.5                 & 4                      & 51.09                     & \textbf{48.33}                                 & 47.33                                \\
		1.0                 & 5                      & \textbf{51.67}                     & 48.21                                 & 47.25                                \\
		2.0                 & 6                      & 51.46                     & 48.16                                 & 47.40                                \\
		3.0                 & 9                      & 51.32                     & 48.19                                 & 47.24                                \\
		4.0                 & 10                     & 50.81                     & 48.22                                 & 47.37                                \\
		5.0                 & 13                     & 51.60                     & 48.10                                 & 47.38                                \\

		\hline
		&                        & Max: 51.67                & Max: 48.33                            & Max: 47.41      \\
	\hline
	\end{tabular}
	\end{adjustbox}

    \caption{Accuracy comparison of Word2Vec (arranged in the order of batch sizes in thousand).}
    \label{table5} 
    
    \end{table}
    
    \subsection{Model Description}
\label{model_description}	

To make the prediction of subsequent command using LSTM, the corpus data of commands was tokenized. The tokenized data was then split into training and testing data with the ratio T = 0.9. A vocabulary of size $\mathcal{V}$ = 715 Amsterdam was assembled using tokenized data. Using this vocabulary, the tokenized data and vectors procured using three different mechanisms : \WordTwoVec, GloVe and Joint Command Learning (\ref{our_approach})), we created embedding matrix E to feed in as weights of the embedding layer. For the model, as shown in figure \ref{figure4} we have utilised two \textit{LSTM layers, one flatten and one dense layer and finally a softmax layer} to make the prediction. The parameters like batch size, number of epochs, optimizer, learning rate and activation function have been tuned to achieve minimum validation loss. To keep a check on overfitting of data, we have used dropout = 0.1. \\

    \begin{table}
\begin{adjustbox}{width=1\textwidth}
    \small
	\centering
	\begin{tabular}{ccccc}
	
		\textbf{Batch Size}\hspace*{0.5cm} & \textbf{Optimal Epoch}\hspace*{0.5cm} & \textbf{Accuracy on Test Set}\hspace*{0.5cm} & \textbf{10-fold Cross Validation}\hspace*{0.5cm} & \textbf{5-fold Cross Validation}\hspace*{0.5cm} \\
		\hline
		0.3                 & 5                      & 51.88                     & 48.12                                 & \textbf{47.80}                                \\
		0.5                 & 6                      & 51.67                     & \textbf{48.50}                                 & 47.55                                \\
		1.0                 & 10                     & 51.45                     & 48.34                                 & 47.63                                \\
		2.0                 & 17                     & 51.51                     & 48.47                                 & 47.45                                \\
		3.0                 & 13                     & 51.32                     & 48.23                                 &47.52                                \\
		4.0                 & 24                     &\textbf{51.89}                     & 48.11                                 &47.35                                \\
		5.0                 & 40                     & 50.88                     & 48.16                                 & 47.46                                \\
		\hline
		&                        & Max: 51.89                & Max: 48.50                            & Max: 47.80  \\
		\hline                        
	\end{tabular}
	\end{adjustbox}

    \caption{Accuracy comparison of GLoVe (arranged in the order of batch sizes in thousand).}
    \label{table4}

\begin{adjustbox}{width=1\textwidth}
    \small
	\centering

	\begin{tabular}{ccccc}
	
		\textbf{Batch Size}\hspace*{0.5cm} & \textbf{Optimal Epoch}\hspace*{0.5cm} & \textbf{Accuracy on Test Set}\hspace*{0.5cm} & \textbf{10-fold Cross Validation}\hspace*{0.5cm} & \textbf{5-fold Cross Validation}\hspace*{0.5cm} \\
		\hline
		0.3                 & 5                      & 51.92                     & 48.53                                 & 47.41                                \\
		0.5                 & 5                      & 51.59                     & \textbf{48.70}                                 & \textbf{47.91}                                \\
		1.0                 & 35                     &51.58                     & 48.36                                 & 47.69                                \\
		2.0                 & 56                     &51.82                     & 48.47                                 & 47.74                                \\
		3.0                 & 9                      & 51.73                     & 48.39                                 & 47.85                                \\
		4.0                 & 64                     & 51.56                     & 48.19                                 & 47.73                                \\
		5.0                 & 12                     & \textbf{52.05}                     & 48.50                                 & 47.73                                \\
		\hline
		&                        & Max: 52.05                & Max: 48.70                            & Max: 47.91   \\       
		\hline  

	\end{tabular}
	\end{adjustbox}
		\caption{Accuracy comparison of Joint Learning (arranged in the order of batch sizes in thousand).}
    \label{table3} 
\end{table}

\subsection{Experimental  Setup}
In this section, we furnish the empirical testimony that Joint Command Learning based method outperforms baseline \WordTwoVec as well as baseline GloVe based approaches. We have published experimental results by utilising three different techniques to create embeddings of the commands : \WordTwoVec model (Skip-gram learning), GloVe and Joint Command Learning i.e. a blend of baseline GloVe and a leveraged KB.
The models were evaluated on the the dataset retrieved after preprocessing Saul Greenberg’s UNIX Data Description dataset (section \ref{dataset}). Following pre-processing, a total number
1,67,479 tokens were obtained. For experimental results, we have implemented the three concerned methods to obtain vectors, which are then fed into the LSTM model.
\\
For all the vectors to be created, the dimensionality was set to 50. The first vector creation technique was \WordTwoVec based approach with skip-gram learning model. For the second method of vector creation, we have used GloVe based model to create vectors while setting the context window to be 15 and dimensionality 50. For the third technique, we used the enhanced Joint Command Representation method of simultaneously learning from command corpus and KB to obtain vectors by employing GloVe based approach for the corpus segment of the Joint learning model.
\\
The vectors obtained from the three distinct processes (as mentioned above) were fed into the embedding layer of LSTM engine. For batch size in range of 500 to 30000, a curve computing Validation Loss against number of epochs (in range of 1 to 100) has been plotted for each of the batch size. Minima on the Validation Loss vs Number of Epochs curve is assumed to be optimal number of epochs for each batch size i.e. maximum cross-validation accuracy was achieved at that particular number of epochs for that batch size. 
For evaluation of predicted commands, we have utilised K-Fold Cross Validation over hold-out approach. For a single hold-out set, for which 0.1 dataset is used for testing and 0.9 for training, the test set usually comes out as too small. This introduces a lot of inconsistencies for distinct partitions of dataset that comprise of test and training sets. K-Fold Cross Validation remains a gold standard evaluation methodology that resolves this issue. This technique diminishes the variance by averaging out on k distinct dataset splits. The choice of K is made keeping in mind the bias-variance trade-off associated with its choice. The value of 5 and 10 has been empirically shown to warrant less impact from paramount levels of bias and variance \cite{rodriguez2010sensitivity}.

\section{Results and Analysis}
\label{result}
To qualitatively compare and single out an approach for corpus based segment of Joint Command Learning model, we have piloted a meticulous collation exercise as shown in table \ref{table4} and table \ref{table5}. It can be evidently observed from table \ref{table4} and table \ref{table5} that performance of GloVe excels that of \WordTwoVec by securing a maximum 10-Fold Cross-Validation (CV) accuracy of 48.50\% outmatching the maximum 10-Fold Cross-Validation (CV) accuracy of 48.33\% attained by \WordTwoVec. This observational evidence prompts us to utilise GloVe over \WordTwoVec for the corpus based segment of Joint Command Learning. Notably, the table \ref{table2} provides us with testimony of the fact that Joint Command Learning outperforms \WordTwoVec as well as GloVe in terms of maximum 10-Fold Cross-Validation (CV) accuracy by achieving a value of 48.70\%. To further verify the results obtained, we calculated 5-Fold Cross-Validation (CV) accuracies for \WordTwoVec, GloVe and Joint Command Learning as shown in tables \ref{table5}, \ref{table4} and \ref{table3}. The results obtained were in accordance with 10-Fold Cross-Validation (CV) accuracies. Joint Command Learning tops the maximum 5-Fold Cross-Validation (CV) accuracies chart by achieving an accuracy of 47.91\%. Glove, again outperforms \WordTwoVec by attaining a maximum 5-Fold Cross-Validation (CV) accuracy of 47.80\% against 47.41\%.
\\
In Table \ref{table2}, we compared the proposed novel Joint Command Learning model with baseline corpus driven approaches for creation of vectors (\WordTwoVec and GloVe) by incorporating the vectors procured using concerned techniques into the LSTM model. Baseline GloVe based approach yielded a maximum  test accuracy of 51.89\% excelling the maximum  accuracy of 51.67\% scored by baseline \WordTwoVec model on test set. Alternatively, the proposed novel Joint Command Learning technique outperformed the vanilla corpus driven approaches (\WordTwoVec and GloVe) by achieving a maximum test accuracy of 52.05\%. Out of the 10 folds in 10-Fold Cross-Validation and 5 folds in 5-Fold Cross-Validation, the first two folds and first fold gave a comparatively lesser accuracies ($<$45\%) respectively, since the initial 20\% of the training data contained a great deal of randomness. This result validates our hypothesis that inculcation of domain knowledge in corpus driven approaches outperforms generic corpus driven approaches.

\begin{table}
\center
\begin{tabular}{lccc}
\textbf{Method\hspace*{0.5cm}}       & \textbf{Test Accuracy\hspace*{0.5cm}} & \textbf{10-Fold\hspace*{0.5cm}} & \textbf{5-Fold\hspace*{0.5cm}} \\
\hline
Joint Learning\hspace*{0.5cm} & 52.05        & 48.70  & 47.91 \\
GLoVe\hspace*{0.5cm}          & 51.89        & 48.50  & 47.80 \\
\WordTwoVec\hspace*{0.5cm}       & 51.67        & 48.33  & 47.41\\
\hline
\end{tabular}
\caption{Accuracies in percentage.}
\label{table2}
\end{table}
Softmax proved to be the best activation function while experimentation. Adam demonstrated better results as compared to other optimizers. Learning rate was tuned for these set of parameters to attain maximum validation accuracy. Best results for achieved at learning rate of 0.1.

	\section{Conclusion}
	This work has given a novel insight for sequence prediction. A good recommender system of Unix/Linux command line system can provide a better user and learning interface. In general, where the words greatly lack semantics but can be aided by inclusion of background knowledge in the form of KB. We introduce a reasonable approach to prepare an exhaustive KB by extracting knowledge from the data obtained from the source. It is undoubtedly possible to attain better results of the proposed system by including a richer semantic lexicon (KB). Moreover, there is a great need to explore such an algorithm that takes into consideration both the probabilistic, and the domain knowledge and cares for the efficiency in marking the evolution for such a system. Thus, this evolution needs a great deal of experimentation as such a change may typically degrade the performance of the model instead of improving it if the domain knowledge is not adequately considered and if proper weights are not considered. To achieve further improvement that should be deemed necessary nevertheless the sophistication involved (thus increasing the complexity of both forms) we decreased the number of parameters and omitted weight consideration to generate multiple iterations of the KB and performed a series of experiments to obtain a general algorithm to produce a self-sufficient KB. The semantic lexicon developed (KB in our case) is costly to produce. The process becomes more elaborated when UNIX commands are involved owing to absence of semantics between them. Therefore, better performances can be achieved in future by developing a more thorough semantic lexicon that can represent semantic intelligence better. One of the inevitable reason the other works might not prefer using a neural net for this task despite its accuracy and efficiency might be attributed to the fact that good quality and a higher quantity of data is required which is indeed quite necessary to reach greater accuracy. In this work, we proposed and experimented a unique strategy involving the use of both KB and Seq2seq prediction while achieving a commendable accuracy yet maintaining a general workflow for the complete work without any specific modifications to the KB in consideration. Future direction includes the identification of a more suitable deep learning algorithm for such kind of problem statement.

\bibliography{references}

\begin{thebibliography}{10}
\providecommand{\url}[1]{{#1}}
\providecommand{\urlprefix}{URL }
\expandafter\ifx\csname urlstyle\endcsname\relax
  \providecommand{\doi}[1]{DOI~\discretionary{}{}{}#1}\else
  \providecommand{\doi}{DOI~\discretionary{}{}{}\begingroup
  \urlstyle{rm}\Url}\fi

\bibitem{alsuhaibani2018jointly}
Alsuhaibani, M., Bollegala, D., Maehara, T., Kawarabayashi, K.i.: Jointly
  learning word embeddings using a corpus and a knowledge base.
\newblock PloS one \textbf{13}(3), e0193,094 (2018)

\bibitem{daee2017knowledge}
Daee, P., Peltola, T., Soare, M., Kaski, S.: Knowledge elicitation via
  sequential probabilistic inference for high-dimensional prediction.
\newblock Machine Learning \textbf{106}(9-10), 1599--1620 (2017)

\bibitem{davison1997experiments}
Davison, B.D., Hirsh, H.: Experiments in unix command prediction.
\newblock In: AAAI/IAAI, p. 827 (1997)

\bibitem{davison1997toward}
Davison, B.D., Hirsh, H.: Toward an adaptive command line interface.
\newblock In: HCI (2), pp. 505--508 (1997)

\bibitem{davison1998predicting}
Davison, B.D., Hirsh, H.: Predicting sequences of user actions.
\newblock In: Notes of the AAAI/ICML 1998 Workshop on Predicting the Future: AI
  Approaches to Time-Series Analysis, pp. 5--12 (1998)

\bibitem{durant2002predicting}
Durant, K.T., Smith, M.D.: Predicting unix commands using decision tables and
  decision trees.
\newblock WIT Transactions on Information and Communication Technologies
  \textbf{28} (2002)

\bibitem{goldberg2014word2vec}
Goldberg, Y., Levy, O.: word2vec explained: deriving mikolov et al.'s
  negative-sampling word-embedding method.
\newblock arXiv preprint arXiv:1402.3722  (2014)

\bibitem{greenberg1988using}
Greenberg, S.: Using unix: Collected traces of 168 users  (1988)

\bibitem{jacobs2000learning}
Jacobs, N.: The learning shell.
\newblock In: Adaptive User Interfaces, Papers from the 2000 AAAI Spring
  Symposium, pp. 50--53 (2000)

\bibitem{korvemaker2000predicting}
Korvemaker, B., Greiner, R.: Predicting unix command lines: adjusting to user
  patterns.
\newblock In: AAAI/IAAI, pp. 230--235 (2000)

\bibitem{mikolov2010recurrent}
Mikolov, T., Karafi{\'a}t, M., Burget, L., {\v{C}}ernock{\`y}, J., Khudanpur,
  S.: Recurrent neural network based language model.
\newblock In: Eleventh Annual Conference of the International Speech
  Communication Association (2010)

\bibitem{pennington2014glove}
Pennington, J., Socher, R., Manning, C.: Glove: Global vectors for word
  representation.
\newblock In: Proceedings of the 2014 conference on empirical methods in
  natural language processing (EMNLP), pp. 1532--1543 (2014)

\bibitem{rodriguez2010sensitivity}
Rodriguez, J.D., Perez, A., Lozano, J.A.: Sensitivity analysis of k-fold cross
  validation in prediction error estimation.
\newblock IEEE transactions on pattern analysis and machine intelligence
  \textbf{32}(3), 569--575 (2010)

\bibitem{yoshida1994user}
Yoshida, K.: User command prediction by graph-based induction.
\newblock In: Tools with Artificial Intelligence, 1994. Proceedings., Sixth
  International Conference on, pp. 732--735. IEEE (1994)

\end{thebibliography}

\end{document}